\documentclass[10pt,twocolumn,letterpaper]{article}

\usepackage[pagenumbers]{cvpr} 

\usepackage{graphicx}
\usepackage{amsmath}
\usepackage{amssymb}

\usepackage{svg}

\usepackage[pagebackref,breaklinks,colorlinks]{hyperref}

\usepackage[capitalize]{cleveref}
\crefname{section}{Sec.}{Secs.}
\Crefname{section}{Section}{Sections}
\Crefname{table}{Table}{Tables}
\crefname{table}{Tab.}{Tabs.}

\def\cvprPaperID{*****} 
\def\confName{NCCV}
\def\confYear{2023}

\begin{document}

\title{Exploring Different Levels of Supervision for Detecting and Localizing Solar Panels on Remote Sensing Imagery}

\author{Maarten Burger\\
Universiteit van Amsterdam\\
Amsterdam, NL\\
{\tt\small maarten.l.burger@gmail.com}
\and
Rob Wijnhoven\\
Spotr.ai\\
Den Haag, NL\\
{\tt\small rob@spotr.ai}
\and
Shaodi You\\
Universiteit van Amsterdam\\
Amsterdam, NL\\
{\tt\small s.you@uva.nl}
}

\maketitle
\pagestyle{plain}
\thispagestyle{plain}

\begin{abstract}
    This study investigates object presence detection and localization in remote sensing imagery, focusing on solar panel recognition. We explore different levels of supervision, evaluating three models: a fully supervised object detector, a weakly supervised image classifier with CAM-based localization, and a minimally supervised anomaly detector. The classifier excels in binary presence detection (0.79 F1-score), while the object detector (0.72) offers precise localization. The anomaly detector requires more data for viable performance.
    Fusion of model results shows potential accuracy gains. CAM impacts localization modestly, with GradCAM, GradCAM++, and HiResCAM yielding superior results. Notably, the classifier remains robust with less data, in contrast to the object detector.
\end{abstract}

\section{Introduction}
\label{sec:intro}

Energy consumption is rising globally~\cite{iea_electricity_report_2023}, and pressing sustainability goals enforce the improvement of energy efficiency of the built environment~\cite{iea_energy_efficiency}. Governments and private corporations are required to update their building portfolios by making them more sustainable. In addition, cities are being made more green by adding trees and vegetation. All such improvements start by measuring the current status as to know on which buildings and areas to focus. The current process typically involves manual inspection of buildings and requires inspectors to physically visit the properties. Remote sensing offers a scalable solution as image data is available at large scale and is frequently being updated. 

While satellite image data has become widely available through modern remote sensing technology, extracting the right information from the images is a challenging process. Object recognition algorithms provide a solution as they can be trained to localize specific objects in the imagery. Although recognition technology has matured over the last decade, datasets to train these systems are only sparsely available and are typically of limited size and quality. In order to obtain high-quality recognition results, it is required to construct specific datasets by collecting and labelling a large number of images. Because labelling is a time-consuming task~\cite{rsi_survey, labelcost}, the question arises how much labelling effort one should spend to obtain a high-quality recognition result. 

In this paper, we evaluate the effect of the level of detail in annotations (supervision) on the resulting recognition accuracy.
Although there has been a lot of work on semi-supervised, weakly supervised and unsupervised methods on popular datasets such as COCO \cite{coco} and ImageNet \cite{imagenet},
very little work has been done on remote sensing imagery~\cite{rsiweak}.
We compare three models with different levels of supervision.
First, a fully-supervised object detector is trained with presence and location information (object boxes).
Secondly, only binary presence labels are used to train a weakly-supervised classification model.
Lastly, we consider a weakly-supervised anomaly detection model that is not trained with any labels, but is only shown images without object of interest.
\begin{figure}[!t]%
    \centering
    \subfloat{{\includegraphics[width=0.3\linewidth]{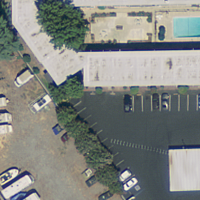} }}%
    \subfloat{{\includegraphics[width=0.3\linewidth]{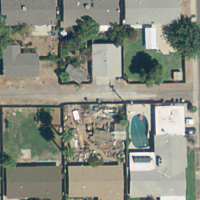} }}%
    \subfloat{{\includegraphics[width=0.3\linewidth]{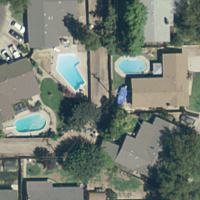} }}%
    \caption{Examples of images in the remote sensing imagery domain with a lot of distractors. There are many visible properties in these images: while the buildings are clear, there are a lot of objects around that might not be of interest.}%
    \label{fig:intro_examples}%
\end{figure}

In our exploration, we experimentally validate these three approaches on the problem of solar panel detection and localization.
Specifically, we investigate the effect of label detail on both the detection (presence) and localization accuracy.
Because classification methods only perform binary detection (no localization), we investigate how much localization information can be extracted by using Class Activation Mapping (CAM) methods such as GradCAM~\cite{gradcam}.
Note that our goal is not to obtain the best-performing models for detection and localization, we use proven baseline models and focus on their comparison.

The main contributions of the paper are:
\begin{itemize}
    \item We compare the accuracy of both binary presence detection and localization for three recognition systems with different levels of supervision: image classification, object detection and anomaly detection,
    \item We show that the classification system provides the best accuracy on presence detection but is worse in localization.
    \item We evaluate the effect of the training data size on the performance and show that the classification system can learn best with limited data.
\end{itemize}

\section{Related work}

We evaluate existing work for image classification, object detection and anomaly detection.

\textbf{Image classification} has long been implemented as a two-stage approach, with fixed filter banks and a trainable classification layer. 
A large increase in accuracy was obtained by Krizhevsky~\cite{krizhevsky2012imagenet}, who proposed the `Alexnet' Convolutional Neural Network (CNN). 
Since then, many different classification networks have been proposed.
One of the seminal works is ResNet by He~\etal~\cite{he2016resnet}, which proposes residual blocks to enable efficient training of really deep networks.
We use a ResNet classification network as it is a widespread baseline for image classification.

The field of \textbf{object detection} has long been a staple of modern computer vision~\cite{object} and focuses on detecting and locating objects in an image.
Numerous approaches have been proposed to tackle this task, each with its strengths and limitations.
Initial CNN-based models such as R-CNN by Girshick~\etal~\cite{rcnn} and Faster R-CNN by Ren~\etal~\cite{fasterrcnn} use a two-stage approach, in which they first generate region proposals and then classify each proposal.
One-stage detectors, such as YOLO by Redmon~\etal~\cite{yolo}, directly predict object class probabilities and boxes in a single pass.
In this work, we use Faster R-CNN as the object detector, as it is widely accepted as a baseline and it is easy to train.

\textbf{Class Activation Mapping (CAM) methods} such as GradCAM~\cite{gradcam} aim to visualize network output decisions in the input image, for recognition models without explicit localization, such as image classifiers.
They have gained popularity in the field of Explainable AI, as these methods are able to extract information from a model that would otherwise not be humanly interpretable. 
Their use and effectiveness as explanations of model output has been both praised \cite{expelephant} and disputed \cite{expstop}.
Several extensions to GradCAM have been proposed, such as GradCAM++~\cite{gradcam++}, HiResCAM~\cite{hirescam}, FullGrad~\cite{fullgrad}, EigenCAM~\cite{eigencam} and EigenGradCAM~\cite{pytorchcam}.
We evaluate the effectiveness of all these methods.

\textbf{Weak supervision in remote sensing} is mainly constrained to object detection~\cite{weakremotesensingsurvey}.
Zhang~\etal~\cite{weakremotesensing3} propose a classification method to automatically gather samples to train an object detector in an unsupervised way.
Qian~\etal~\cite{wsod_semantic_segmentation_2023} extend this idea to weakly supervised semantic segmentation by addressing issues related to unreliable pseudo ground truth instances and inaccurate object localization.
In contrast to specific solutions, we seek to compare different and relatively simple existing methods.

\textbf{Anomaly detection}.
Anomaly detection considers implicit recognition of the objects of interest by only learning from images without these objects, so without any class labels.
The variational autoencoder (VAE)~\cite{vae} is a model originally built for being able to learn intractable posterior distributions, which enables it to be used for anomaly detection~\cite{avae1} as shown in~\cite{vae_anomaly}.
The input data is compared to the reconstructed counterpart, where anomalies are then detected by way of the reconstruction error. However, accurately localizing these anomalies requires finding a threshold which actually requires the abnormal images. 
Note that Generative Adversarial Networks (GANs) have also been used for this task~\cite{gan5} but are hard to train.
Therefore, we focus on the VAE in our exploration.

\section{Methods}

\begin{figure}[!t]%
    \centering
    \includesvg[inkscapelatex=false,width=1\linewidth]{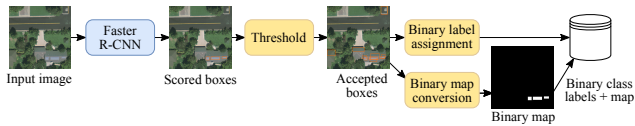}
    \caption{Object detector pipeline.}
    \label{fig:object_detector_diagram}
\end{figure}

We now describe the three models that use different levels of supervison and provide implementation details.
First, the fully-supervised object detection system, which uses multiple box labels per image. 
Next, the classification method that evaluates object presence using weakly-supervised image-level labels. 
Finally, the anomaly detector is implemented as a Variational Auto Encoder (VAE), which uses even more weak supervision by only exploiting images without objects (no positive samples).
Note that all three models are trained from scratch (no pre-trained weights), to enable an objective comparison.
\\
\\
\noindent\textbf{Fully-supervised object detection} is implemented using the Faster R-CNN detector by Ren~\etal~\cite{fasterrcnn}.
We train it on the boxes from the dataset and threshold the detections.
Resulting detections are converted into a binary pixel mask for the evaluation of the localization performance.
An overview of the pipeline can be found in Figure \ref{fig:object_detector_diagram}.
For the experiments, we use an existing PyTorch implementation~\cite{pytorch_faster_rcnn}.
The model is trained using conventional methods: utilizing the Adam optimizer~\cite{adam}, weighted Cross Entropy loss for the classification loss and Smooth L1-Loss for the bounding box loss.
\\
\\
\noindent\textbf{Weakly-supervised classification} uses a ResNet-50 network~\cite{resnet} with a classification head.
For comparison reasons, we use the same backbone architecture.
Because classification only results in a single presence score, we add a CAM method to generate a heatmap that explains the classification outcome. 
The generated class activation map from the CAM model (the heatmap) is thresholded to obtain a binary localization map.
We use the PyTorch implementation of ResNet-50~\cite{pytorch_resnet_50} with a separate head for binary classification. 
Similar to the object detector, we do not use pre-trained weights and train from scratch, utilizing Adam optimization and a weighted binary Cross Entropy loss.
\begin{figure}[!t]%
    \centering
    \includesvg[inkscapelatex=false,width=1\linewidth]{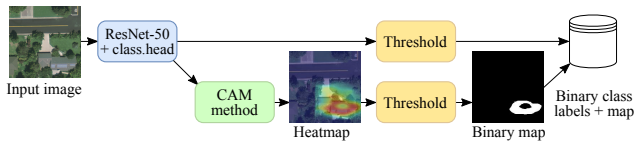}
    \caption{Classification pipeline.}
    \label{fig:classification_diagram}
\end{figure}

\begin{figure}[!t]%
    \centering
    \includesvg[inkscapelatex=false,width=1\linewidth]{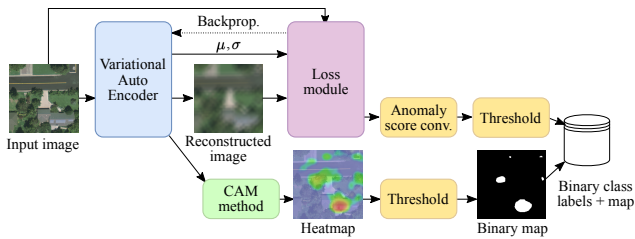}
    \caption{Variational Autoencoder (VAE) pipeline.}
    \label{fig:vae_diagram}
\end{figure}
We experiment with different CAM methods: GradCAM~\cite{gradcam}, GradCAM++~\cite{gradcam++}, FullGrad~\cite{fullgrad}, EigenCAM~\cite{eigencam}, EigenGradCAM~\cite{pytorchcam} and HiResCAM~\cite{hirescam}.
For all of these methods, we used PyTorch modules as implemented by~\cite{pytorchcam}.
An overview of the pipeline can be found in Figure~\ref{fig:classification_diagram}. 
\\
\\
\noindent\textbf{Weakly-supervised variational autoencoder (VAE)} is trained to learn what a `normal' images looks like.
High reconstruction loss is expected when observing `abnormal' images (anomalies).
The VAE is manually implemented to follow a similar structure to ResNet-50 in order to ensure a fair comparison against both the object detector and the classifier. 
The encoder follows the ResNet-50 architecture, but instead of a single classification head there are two final layers, which calculate the distribution parameters: one linear layer that outputs $\mu$ (mean) and one linear layer that outputs $\sigma$ (variance). 
Using these distribution parameters, we sample conventionally using reparameterization. 
The result of this sample ($z$), is then used as input for the decoder part of the VAE. 
The decoder follows a symmetrical structure to the encoder (same operations), but in reverse order and using deconvolutional layers. 
Although typical scaling performs a singular sigmoid operation, we scale the loss based on the minimum and maximum reconstruction losses from the validation set (to range [0-1]).
The size of the latent space is left as a tunable hyperparameter.
The model is not again pre-trained and trained with the Adam optimizer and utilizes the Kullback–Leibler divergence as the latent loss and mean square error as the reproduction loss.
An overview of the pipeline can be found in Figure \ref{fig:vae_diagram}. 

We train the VAE with images of buildings that do not contain solar panels, such that a solar panel should be detected as abnormal and thus an anomaly during inference.
Note that it is expected that other uncommon objects will also be detected as anomalies.
A CAM method is then used to extract a heatmap of where the anomaly is located. 
Although a CAM method requires a target class, we use the original image as the target as to extract the anomaly heatmap. 

\vspace{5mm}
\section{Experiments}
\label{sec:metrics}

We now evaluate each of the three models for presence detection (is there a solar panel present in the image) and localization (where is the solar panel located). 
Each model outputs both image-level classification output and a per-pixel binary localization map, which enables their comparison.

\subsection{Dataset: solar panels}
\begin{figure}[!t]%
    \centering
    \subfloat[\centering ]{{\includegraphics[width=0.33\linewidth]{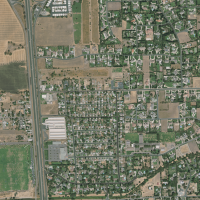} }}%
    \subfloat[\centering ]{{\includegraphics[width=.33\linewidth]{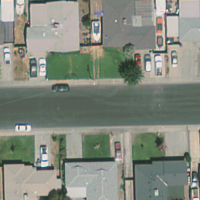} }}%
    \subfloat[\centering ]{{\includegraphics[width=.33\linewidth]{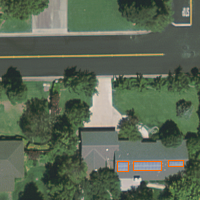} }}%
    \\
    \subfloat[\centering ]{{\includegraphics[width=0.33\linewidth]{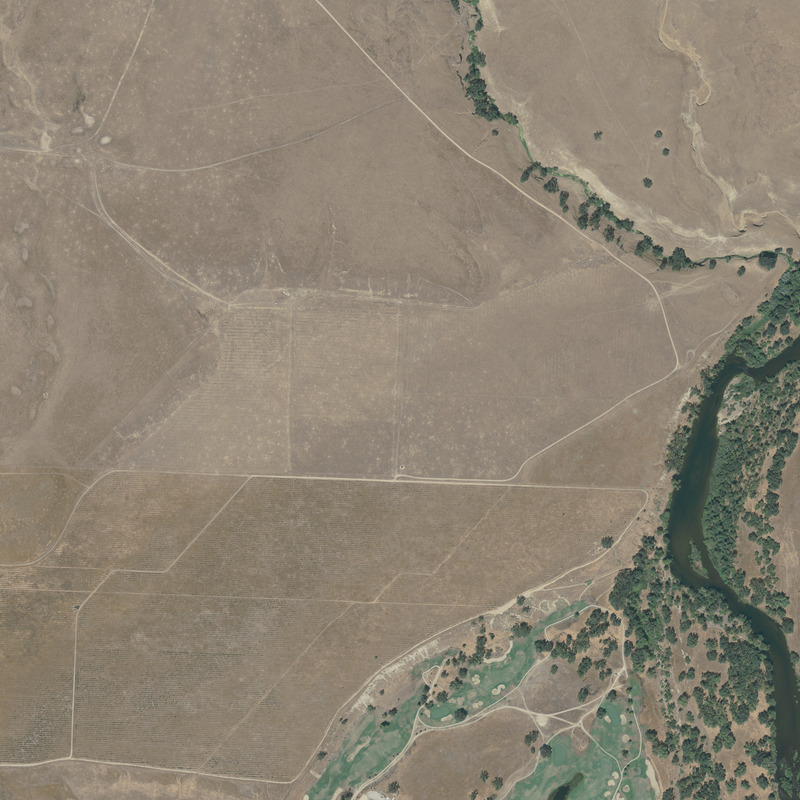} }}%
    \subfloat[\centering ]{{\includegraphics[width=0.33\linewidth]{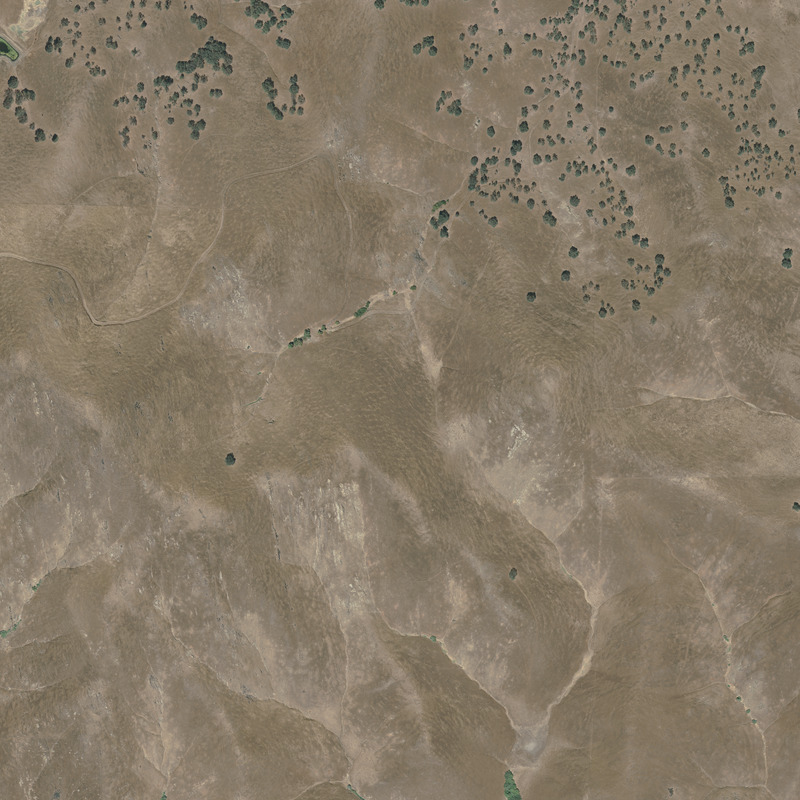} }}%
    \subfloat[\centering ]{{\includegraphics[width=0.33\linewidth]{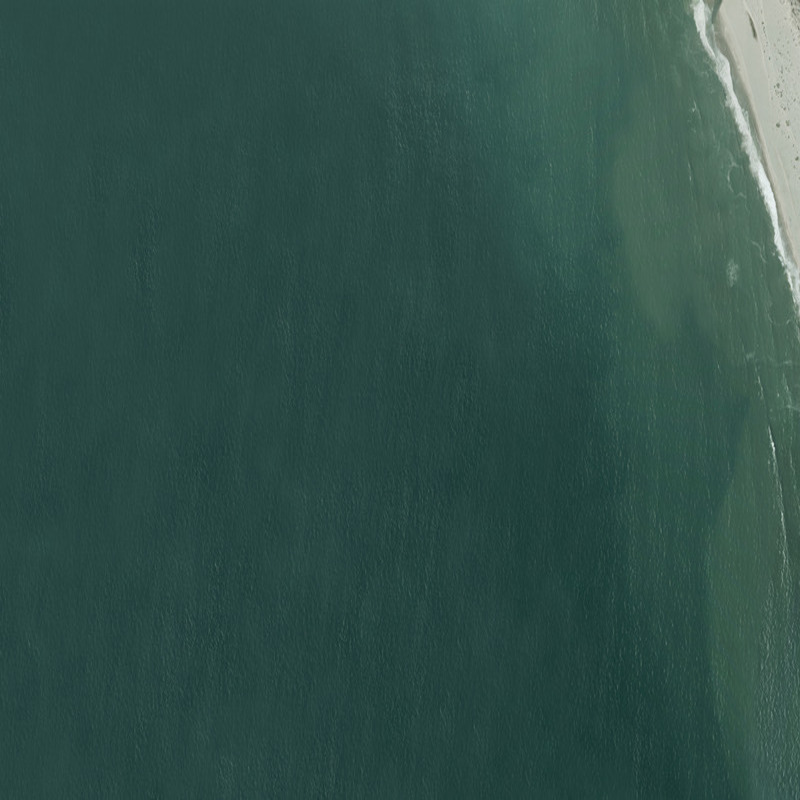} }}%
    \caption{(a) High-resolution image of $5000\times5000$ pixels. (b) Cropped $200\times200$ pixel image tile with no solar panels present. (c) Tile with solar panels (polygons in orange). (d,e,f) Examples of high-res images without panels.}%
    \label{fig:dataset_examples}%
\end{figure}
We use the ``Distributed Solar Photovoltaic Array Location and Extent Data Set for Remote Sensing Object Identification'' dataset~\cite{Dataset}.
This dataset was originally published as~\cite{Datasetoriginal}, but has been extended with more imagery and some faulty annotations have been rectified.
The dataset consists of 601 high quality RGB satellite images (at 0.33 meter per pixel resolution) across four cities in California in the United States of America: Fresno, Stockton, Modesto and Oxnard. 
Each image has a resolution of $5000\times5000$ pixels, encompassing an area of $1.7\times1.7$ kilometers.
Solar panels are not annotated individually, each polygon annotation considers a group of connected panels.
The total dataset comprises 19,821 solar panel polygons.
Note that in contrast to rectangular box annotations, the polygons allow for more accurate evaluation of the localization performance.
We did also evaluate the DeepSolar~\cite{deepsolar} dataset, but although it contains a larger number of image, the images are only annotated with a binary presence label (weak supervision) to indicate the presence of at least one solar panel in the image.
\\
\\
\noindent\textbf{Pre-processing.} 
We pre-process the images before using them in our experiments. 
First, we split each single original high-resolution images into 625, non-overlapping, smaller images of size $200\times200$ pixels ($67\times67$ meters), resulting in a total of 375,625 small images.
Because the total dataset covers large parts of areas without buildings, there is a strong class imbalance as most image tiles do not contain solar panels. 
To avoid significant bias in the model training, we removed 138 high-resolution images that did not contain any solar panels (see Figure~\ref{fig:dataset_examples} d, e, f).
The number of images cleaned this way are displayed in Table \ref{tab:dataset_processing}.
To avoid detection difficulties around borders, we remove small polygons if their size is below five pixels in area or either dimension is below two pixels in size.
Some examples of Figure \ref{fig:dataset_examples}. 
To be able to train the object detector, we convert the polygons to rectangular boxes.
For the image classifier, we convert to binary presence labels per image.
In this work we split the dataset into three conventional splits: a training set, a validation set and a test set. We utilize a conventional 80-10-10 split. 
For the VAE the dataset is adapted slightly. 
Since we use the VAE for anomaly detection, the VAE should only have images without solar panels in the training set. 
So we simply use the same training set as the other methods, but we remove the images with solar panels. 
The validation and test set remain the exact same.

\begin{table}
\centering
\begin{tabular}{c|rrrr}
\hline
                         & Original & Cleaned &  Tiles \\ \hline
Total Images             & 601      & 463     & 289,375 \\
Polygons                 & 19,821   & 19,821  &  20,961 \\
Images with panels       & 463      & 463     &  13,065 \\
Images without panels    & 138      & 0       & 276,310 \\ \hline
\end{tabular}
\caption{Solar panel dataset statistics.}
\label{tab:dataset_processing}
\end{table}

\subsection{Evaluation metrics}

Presence detection results in a single label per image, which is evaluated using F1-score as the main reported metric~\cite{fscore_orig}.
Localization will be evaluated on true positives only, as also performed in explainable AI and object detection.
This means that if the model predicts a positive classification/confidence score that is congruent with the label of the image, localization metrics are calculated. 
We only consider true positives, as it is impossible to evaluate on images that do not contain ground-truth localization and it does not make sense to localize a property which the  model does not believe is exists (negative classification confidence).
For localization we use conventional segmentation metrics, namely the overlap metrics DICE-score (DICE) and Intersection over Union (IoU)~\cite{unet}.
Although DICE and IoU are directly proportional to each other ($DICE \propto IoU$), we include both measures for completeness. 
We also include a percentage for the amount of images where the DICE-score (and IoU) is zero, which we call ``no overlap''.
For these images, there is no overlap between the predicted localization and the ground-truth polygons.

All quantitative results are the mean of three individual runs (so three separately trained models).
If the standard deviation is $< 0.03$, it is unreported.
If the standard deviation is $\geq 0.03$ it is reported in the format: $\mu (\pm \sigma)$ where $\mu$ is the mean across three runs and $\sigma$ is the standard deviation. 

\subsection{Hyperparameters}
\label{sec:hyperparams}
We first perform a hyperparameter search for all our models by evaluating on the validation set.
The best hyperparameters are then selected for the next experiments. 
The results of this search and the search space are included in Table \ref{tab:hyperparams}.

\begin{table}[!t]
\scalebox{0.85}{
\begin{tabular}{c|ccc}
\hline
Hyperparameter             & Detector                      & Classifier                           & VAE                 \\ \hline
Encoder                    & ResNet-50                     & ResNet-50                             & ResNet-50           \\ \hline
Epochs                     & 10                            & 10                                    & 10                            \\
Batch Size                 & 8                             & 14                                    & 5                             \\
Optimizer                  & Adam                          & Adam                                  & Adam                          \\
Learning rate              & 0.0001                        & 0.0001                                & 0.00005                       \\
Solar pnl. loss w          & 20                            & 20                                    & -                             \\
Reconstr. loss w           & -                             & -                                     & 0.9                           \\
Latent dims.               & -                             & -                                     & 4096                          \\ \hline
Hyperparameter             & \multicolumn{3}{c}{Search Space}                                                                      \\ \hline
Epochs                     & \multicolumn{3}{c}{\{1, 3, 5, 8, 10, 20\}}                                                            \\
Batch size                 & \multicolumn{3}{c}{\{1, 2, 3, ..., 23, 24\}}                                                          \\
Optimizer                  & \multicolumn{3}{c}{\{Adam, AdamW, Adagrad, RMSprop, ASGD\}} \\
Learning rate              & \multicolumn{3}{c}{\{0.1, 0.01, 0.005, 0.001, ..., 0.000001\}} \\
Solar pnl. loss w          & \multicolumn{3}{c}{\{1, 5, 10, 20, 30, 50, 100\}}                                                     \\
Reconstr. loss w           & \multicolumn{3}{c}{\{0.0, 0.1, 0.2, 0.3, ... 0.9, 1.0\}}                                              \\
Latent dims.               & \multicolumn{3}{c}{\{32, 64, 128, ..., 8192, 16384\}}                          \\ \hline
\end{tabular}}
\caption{The results of the hyperparameter search.}
\label{tab:hyperparams}
\end{table}

\subsection{Experiment 1: performance and comparison}
\label{sec:comparison}

The first experiment is a quantitative measurement of both detection and localization performance across all models using the metrics we described above.
We also perform a qualitative analysis for the localization by manually inspecting a number of generated bounding boxes and heatmaps.
We use the best performing model for each method to quantitatively compare the methods against each other.

\paragraph{Object detection}

\begin{table}[!t]
\scalebox{0.75}{
\begin{tabular}{c|c|cc|cc|c}
\hline
Thresh.          & Detection      & \multicolumn{2}{c|}{Polygons}   & \multicolumn{2}{c|}{Bounding Boxes} &                 \\
Object Det.      & F1-score        & DICE           & IoU           & DICE            & IoU           & no overlap      \\ \hline
0                & 0.480          & 0.623          & 0.450          & 0.776          & 0.630          & 1.17\%          \\
0.1              & 0.582          & 0.686          & 0.547          & 0.789          & 0.684          & 1.69\%          \\
0.2              & 0.679          & 0.687          & 0.548          & 0.791          & 0.687          & 1.55\%          \\
0.3              & 0.709          & 0.689          & 0.5487          & 0.796          & 0.692          & 1.35\%          \\
0.35             & \textbf{0.720} & \textbf{0.722} & \textbf{0.611} & \textbf{0.810} & \textbf{0.707} & \textbf{1.12\%} \\
0.4              & 0.714          & 0.694          & 0.555          & 0.797          & 0.700          & 1.48\%          \\
0.5              & 0.695          & 0.663          & 0.520          & 0.766          & 0.658          & 1.59\%          \\
0.6              & 0.613          & 0.648          & 0.491          & 0.738          & 0.615          & 2.11\%          \\
0.7              & 0.577          & 0.640          & 0.497          & 0.741          & 0.631          & 2.69\%          \\
0.8              & 0.412          & 0.631          & 0.494          & 0.728          & 0.626          & 6.23\%          \\
0.9              & 0.177          & 0.600          & 0.430          & 0.701          & 0.586          & 13.11\%         \\
0.95             & 0.068          & 0.272          & 0.261          & 0.378          & 0.286          & 65.74\%         \\
0.999            & 0              & 0              & 0              & 0              & 0               & 100\%           \\ \hline
\end{tabular}}
\caption{The localization performance of the object detection model for different thresholds.}
\label{tab:object_detection_localization}
\end{table}

From the results in Table~\ref{tab:object_detection_localization} we observe that the best detection and localization accuracy is obtained at a relatively low threshold value (value 0.35).
This suggests that the model has not been fully converged, which could originate from the relatively small training set size.
Overall there is very little variance across runs as all results except for one ($threshold=0.1$) have a low standard deviation of $<0.03$.
Across all three runs there were no boxes containing a score of $\geq 0.999$.
The highest accuracy (F1-score) is obtained at a threshold of 0.35.
The achieved F1-score of 0.720 will provide a baseline for the other methods.
As expected, the localization accuracy on the bounding boxes as ground-truth is higher than on polygons. 

\begin{figure}[!b]%
    \centering
    \subfloat [\centering] {{\includegraphics[width=0.33\linewidth]{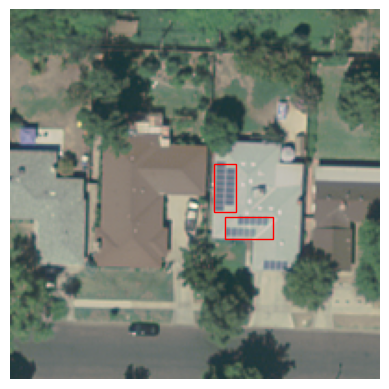} }}%
    \subfloat [\centering] {{\includegraphics[width=0.33\linewidth]{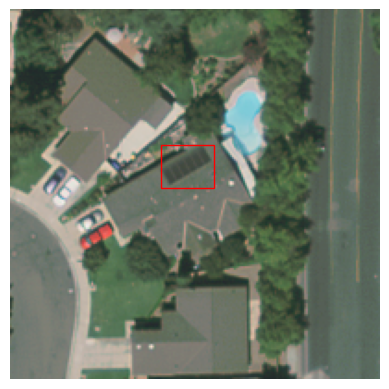} }}%
    \subfloat [\centering] {{\includegraphics[width=0.33\linewidth]{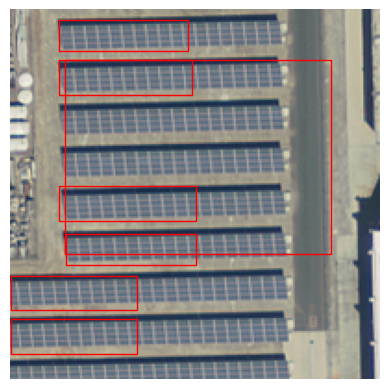} }} \\
    \subfloat [\centering] {{\includegraphics[width=0.33\linewidth]{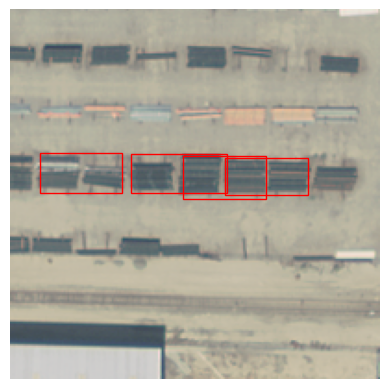} }}%
    \subfloat [\centering] {{\includegraphics[width=0.33\linewidth]{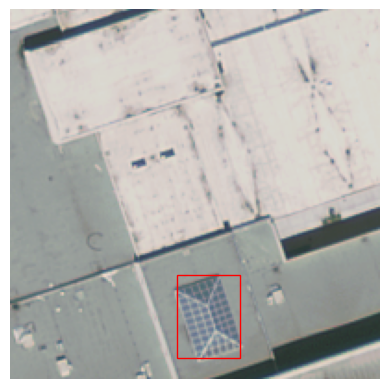} }}%
    \subfloat [\centering] {{\includegraphics[width=0.33\linewidth]{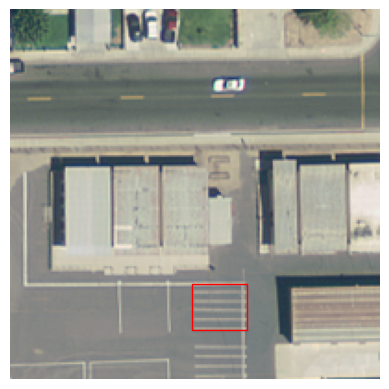} }}%
    \caption{Examples of localization results of the object detector, for correct positive binary labels (a-c) and incorrect labels (d-f) (threshold 0.35).}%
    \label{fig:object_detector_results}%
\end{figure}

For a qualitative analysis we manually inspected 412 images.
We included some typical and a few interesting examples of true positives in Figure~\ref{fig:object_detector_results}.
Most true positive examples have the solar panel bounding boxes at the correct location, as can be seen in Figure~\ref{fig:object_detector_results}~(a) and (b).
Sometimes, when multiple solar panels are present, the model misses a few.
This happens mostly when panels are numerous and tightly packed together (see Figure~\ref{fig:object_detector_results}~(c)).
We expect that this is partly caused by the non-maximum suppression algorithm that merges the initial detection boxes.
Also, the detector performs multi-scale detection, which is not desired for remote sensing imagery of fixed scale.
Most false detections were observed on visually similar objects such as steel beams (Figure~\ref{fig:object_detector_results}~(d)), skylights (e), painted rectangular high-contast rectangular shapes (e).

\paragraph{Classification.}
The detection and localization performance of the classification model with GradCAM is shown in Table~\ref{tab:classification_gradcam_localization}. 
In contrast to the object detector, we observe that the optimal threshold for detection is at 0.95, which hints that the training process has converged more.
Interestingly, the optimal threshold for localization is lower (0.7), showing that the model performs differently for the detection and localization tasks.
The DICE-score (and IoU) varies significantly over the evaluated threshold range.
Therefore, the correct working point should be carefully selected when targeting localization.
There is a slightly higher overall performance when evaluated on the bounding boxes rather than polygons, likely due to the bounding boxes covering a larger overall area which 'absorbs' slight localization errors around the polygon edges.
Also interesting is the values in the ``No overlap'' column at thresholds 0.9, 0.95 and 0.999.
Here it can be observed that for around 85\% of the images, the area with top 10\% activation has direct overlap with a solar panel.
For around 70\% of the images, the area with the top 5\% activation includes has direct overlap with a solar panel.
Lastly, for around 43\% of the images, the area with the top 0.1\% activation includes has direct overlap with a solar panel. 

\paragraph{Effect of different CAM methods.} 
Localization metrics for all evlauated CAM models are shown in Table~\ref{tab:classifier_cam_compilation}.
Note that the reported scores are the ones with the most optimal binarization threshold, resulting in the highest DICE-score on the ground-truth polygons.
For all CAM methods the threshold value of $0.7$ was optimal.
GradCAM++ and HiResCAM are an improvement over GradCAM, while FullGrad, EigenCAM and EigenGradCAM do not outperform.
Overall, the results show that there is not that much difference in performance between the different CAM-methods.
This falls in line with expectations as these CAM-methods all attempt to extract the same type of information from the same model.

\begin{table}[!t]
\scalebox{0.75}{
\begin{tabular}{c|c|cc|cc|c}
\hline
Threshold                    & Detection       & \multicolumn{2}{c|}{Polygons}   & \multicolumn{2}{c|}{Bounding Boxes} &        \\
\multicolumn{1}{l|}{GradCAM} & F1-score         & DICE            & IoU            & DICE           & IoU            & No overlap \\ \hline
0                            & 0.086           & 0.055          & 0.030          & 0.068          & 0.038          & \textbf{0.00\%}     \\
0.05                         & 0.194             & 0.095          & 0.053          & 0.118          & 0.067          & 0.24\%     \\
0.1                          & 0.374           & 0.116          & 0.065          & 0.144          & 0.083          & 0.24\%     \\
0.2                          & 0.451           & 0.157          & 0.090          & 0.192          & 0.114          & 0.24\%     \\
0.3                          & 0.517           & 0.199          & 0.117          & 0.240          & 0.146          & 0.24\%     \\
0.4                          & 0.615           & 0.247          & 0.148          & 0.293          & 0.183          & 0.35\%     \\
0.5                          & 0.644           & 0.300          & 0.185          & 0.348          & 0.224          & 0.47\%     \\
0.6                          & 0.687           & 0.349          & 0.221          & 0.392          & 0.258          & 0.59\%     \\
0.7                          & 0.714           & \textbf{0.377} & \textbf{0.244} & \textbf{0.407} & \textbf{0.271} & 0.71\%     \\
0.8                          & 0.743           & 0.342          & 0.221          & 0.355          & 0.233          & 3.18\%     \\
0.9                          & 0.779           & 0.210          & 0.131          & 0.211          & 0.133          & 15.43\%    \\
0.95                         & \textbf{0.790}  & 0.120          & 0.073          & 0.120          & 0.073          & 29.92\%    \\
0.999                        & 0.669           & 0.027          & 0.017          & 0.026          & 0.016          & 57.14\%    \\ \hline
\end{tabular}}
\caption{The detection and localization performance of the classifier with GradCAM across different thresholds.}
\label{tab:classification_gradcam_localization}
\end{table}

\begin{table}[!t]
\centering
\scalebox{0.75}{
\begin{tabular}{c|cc|cc|c}
\hline
            & \multicolumn{2}{c|}{Polygons}  & \multicolumn{2}{c|}{Bounding Boxes} &     \\
CAM Method  & DICE           & IoU            & DICE           & IoU           & No Overlap \\ \hline
GradCAM     & 0.377          & 0.244          & 0.407          & 0.271         & 0.71\%     \\
GradCAM++   & \textbf{0.390} & \textbf{0.255} & \textbf{0.423} & 0.256         & 0.83\%     \\
FullGrad    & 0.345          & 0.217          & 0.379          & 0.245         & \textbf{0.59\%}     \\
EigenCAM    & 0.335          & 0.218          & 0.358          & 0.238         & 14.11\%    \\
EigenGradCAM & 0.336         & 0.219          & 0.358          & 0.238         & 14.11\%    \\
HiResCAM    & 0.385          & 0.252          & 0.416          &\textbf{0.279}  & 1.56\%     \\ \hline
\end{tabular}}
\caption{Localization accuracy for different CAM-methods. Only GradCAM++ and HiResCAM outperform the GradCAM baseline.}
\label{tab:classifier_cam_compilation}
\end{table}

\begin{figure}[!h]%
    \centering
    \subfloat [\centering] {{\includegraphics[width=0.33\linewidth]{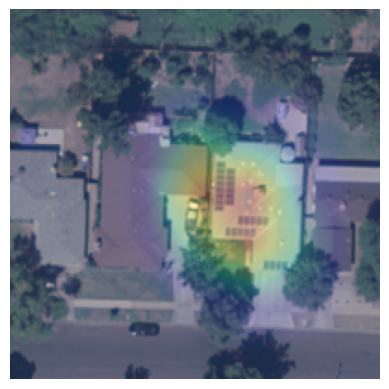} }}%
    \subfloat [\centering] {{\includegraphics[width=0.33\linewidth]{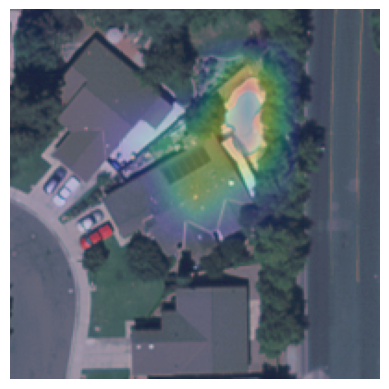} }}%
    \subfloat [\centering] {{\includegraphics[width=0.33\linewidth]{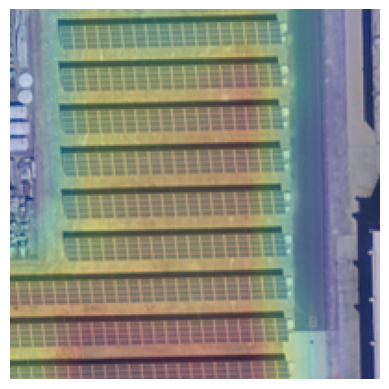} }} \\
    \subfloat [\centering] {{\includegraphics[width=0.33\linewidth]{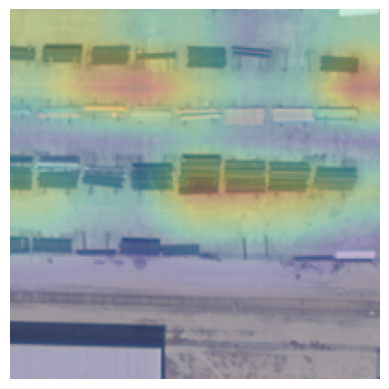} }}%
    \subfloat [\centering] {{\includegraphics[width=0.33\linewidth]{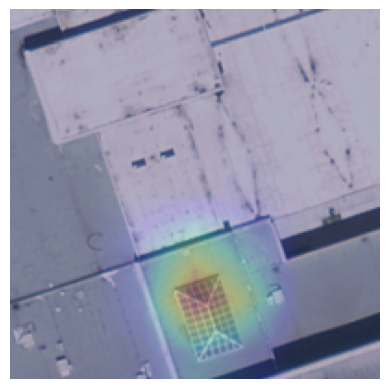} }}%
    \subfloat [\centering] {{\includegraphics[width=0.33\linewidth]{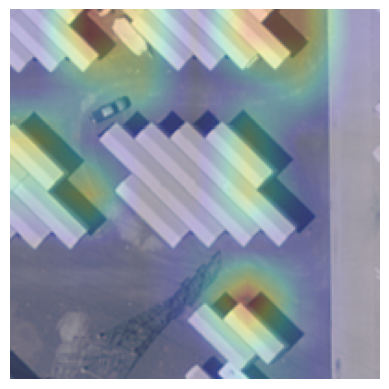} }}%
    \caption{Examples of detection explanations with GradCAM localization for correct (a-c) and incorrect detection (d-f).}%
    \label{fig:classifier_gradcam_tp}%
\end{figure}

\begin{figure}[!h]%
    \centering
    \subfloat [\centering] {{\includegraphics[width=0.33\linewidth]{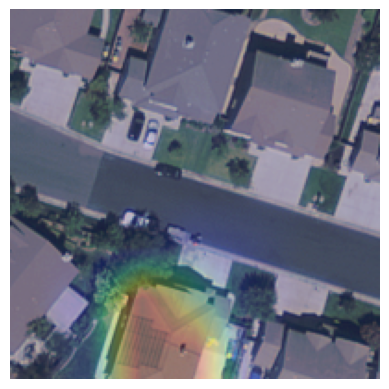} }}%
    \subfloat [\centering] {{\includegraphics[width=0.33\linewidth]{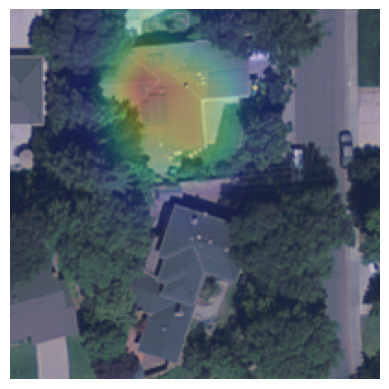} }}%
    \subfloat [\centering] {{\includegraphics[width=0.33\linewidth]{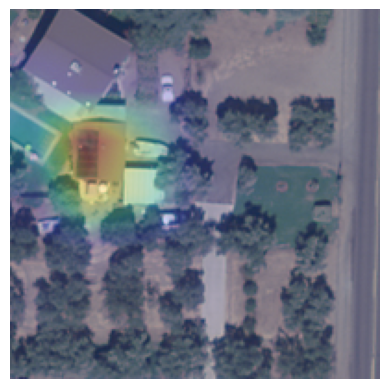} }}%
    \caption{Examples of misclassified images that have high visual resemblance to solar panels and might fool human observers.}%
    \label{fig:classifier_human_fool}
\end{figure}

\begin{figure}[!h]%
    \centering
    \subfloat [\centering GradCAM] {{\includegraphics[width=0.33\linewidth]{imgs/classifier/tp/solar_panel_332_3.png} }}%
    \subfloat [\centering GradCAM++] {{\includegraphics[width=0.33\linewidth]{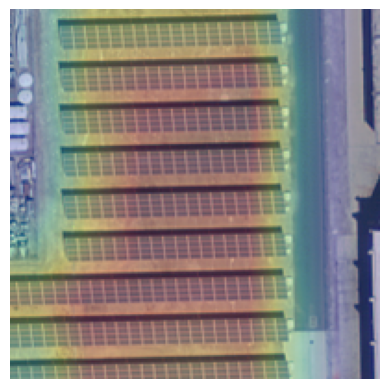} }}%
    \subfloat [\centering FullGrad] {{\includegraphics[width=0.33\linewidth]{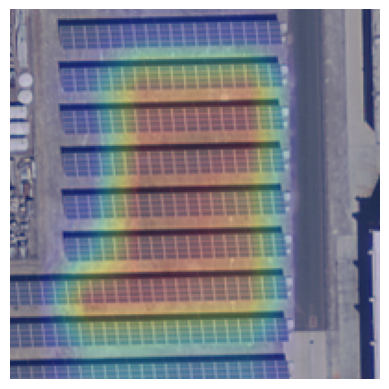} }} \\
    \subfloat [\centering EigenCAM] {{\includegraphics[width=0.33\linewidth]{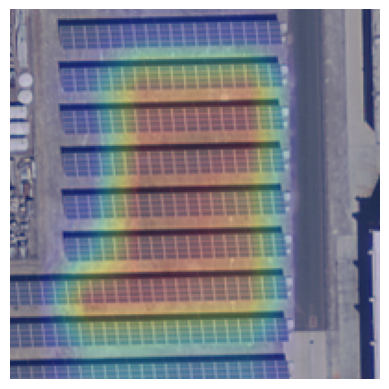} }}%
    \subfloat [\centering EigenGradCAM] {{\includegraphics[width=0.33\linewidth]{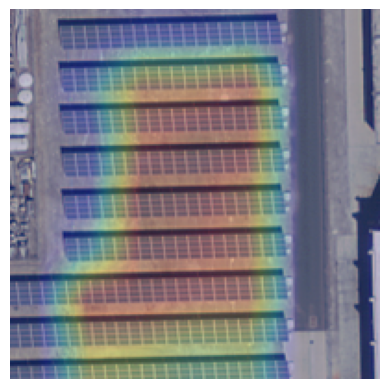} }}%
    \subfloat [\centering HiResCAM] {{\includegraphics[width=0.33\linewidth]{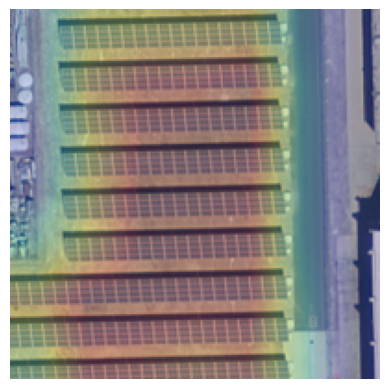} }}%
    \caption{Examples of different CAM methods applied on an image with a large number of solar panels.}%
    \label{fig:classification_cam_compilation}%
\end{figure}

During the quantitative analysis, we noticed that across all CAM methods, the models focused on the same general areas.
Figure~\ref{fig:classification_cam_compilation} shows different CAM activations for an image with a large number of solar panels.
As can be seen in the figure, the heatmaps are overall quite similar and the main difference is how the activation heatmaps spread from the peak activation (in red). 

Overall, we found very few images where the heatmap did not have high activation around a solar panel.
This finding is supported by the “No overlap” values of Table~\ref{tab:classification_gradcam_localization}, where the values are very low (even at higher thresholds), suggesting that high activation on a solar panel is present in nearly all images.
We did an interesting observation, namely the behaviour of CAM-based localization in images with a high number of solar panels that are not all directly attached.
Generally there was very high activation around a subset of these solar panels, but not always on all panels.
This is likely due to the nature of the classification task, where finding one instance of a solar panel already constitutes as assigning a positive classification label.
So after this, the model already has the information it needs to make a prediction and does not require more information from the image.
Examples shown in Figure~\ref{fig:classifier_gradcam_tp}~(d) and (f).

Among the false positives there were a few properties that stood out as particularly common: steel beams, rectangular shadows, skylights and rectangular roof objects in general. Images with examples of these properties can be seen in Figure \ref{fig:classifier_gradcam_tp}~(d-f). For some false positives we even started doubting if the ground-truth labels are incorrect, as some human observers might be make the same false decision (see Figure \ref{fig:classifier_human_fool}).

\paragraph{Anomaly Detection}

The detection and localization performance of the anomaly detection model is found in Table~\ref{tab:vae_localization}. 
In general, the anomaly detection model has very poor detection and localization performance.
We expect that this originates from the more complex problem that it tries to solve, as no explicit positive samples have been observed by the model.
Overall, the localization performance is again slightly higher when evaluated on the bounding boxes rather than polygons, likely due to the bounding boxes covering a slightly larger area.
In 26\% of all images there is no overlap with any solar panel, indicating the detection of another anomaly in these images.

\begin{figure}[!t]%
    \centering
    \subfloat [\centering] {{\includegraphics[width=0.33\linewidth]{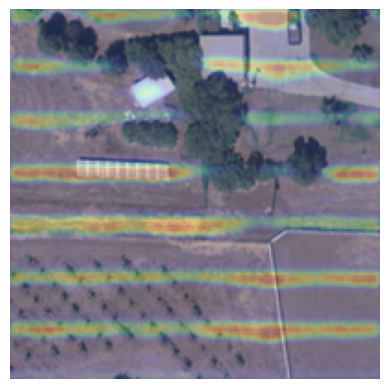} }}%
    \subfloat [\centering] {{\includegraphics[width=0.33\linewidth]{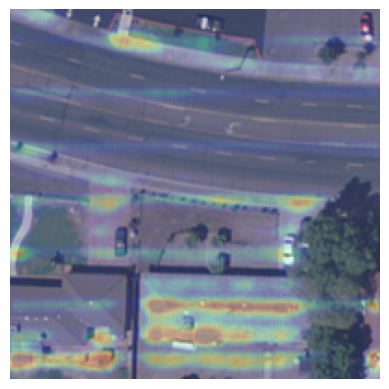} }}%
    \subfloat [\centering] {{\includegraphics[width=0.33\linewidth]{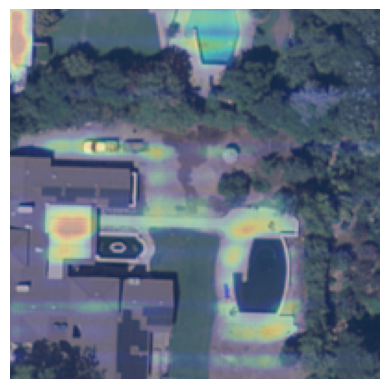} }} \\
    \subfloat [\centering] {{\includegraphics[width=0.33\linewidth]{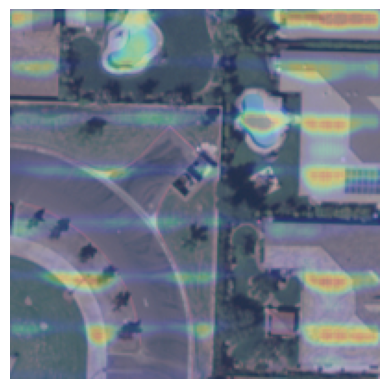} }}%
    \subfloat [\centering] {{\includegraphics[width=0.33\linewidth]{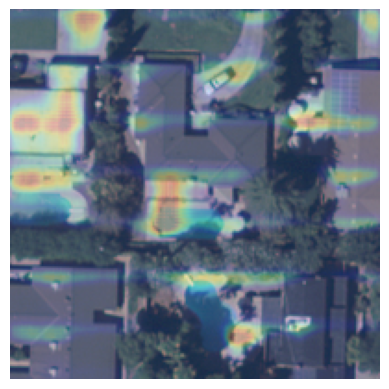} }}%
    \subfloat [\centering] {{\includegraphics[width=0.33\linewidth]{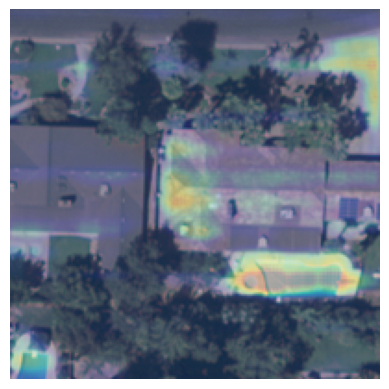} }}%
    \caption{Examples of anomaly detection maps on images containing solar panels (a-c) and detections on swimming pools (d-f).}%
    \label{fig:vae_tp}%
\end{figure}

During the quantitative analysis, the first thing we noticed were the fractured and unfocused anomaly heatmaps, where it seems to be a compilation of multiple heatmaps instead of focusing on a single area.
Also some striping artifacts appear, which we cannot yet fully explain.
In Figure~\ref{fig:vae_tp}~(a) and (b), there is overlap with the solar panels, but in (c) there is seemingly no overlap at all, instead mainly focusing on a white building between the solar panels and the swimming pool at the top of the image, albeit to a lesser extent.

Further in this quantitative analysis we noticed that there was a large amount of images with swimming pools present among the detected anomalies. We have included some examples in Figure~\ref{fig:vae_tp}~(d-f), where we also see a high heatmap activation around the swimming pool area. 
Interesting to see is that a bright blue swimming pool generally has a lot of heatmap activation but a darker blue swimming pool gets much less attention from the anomaly localizer (Figure~\ref{fig:vae_tp}~(f)).
During the qualitative analysis, we have seen many incorrect activations on trucks, swimming pools and tennis fields.
Essentially, anything that is not a normal building, a road or a brown/green field seems to be a possible anomaly.

\begin{table}[!t]
\scalebox{0.75}{
\begin{tabular}{c|c|cc|cc|c}
\hline
Threshold & Detection      & \multicolumn{2}{c|}{Polygons}                                     & \multicolumn{2}{c|}{Bounding Boxes}                               &              \\
VAE       & F1-score        & DICE           & IoU            & DICE           & IoU            & No overlap   \\ \hline
0         & 0.026          & 0.052          & 0.030          & 0.054          & 0.031          & \textbf{0\%} \\
0.1       & 0.062          & 0.080          & 0.047          & 0.086          & 0.050          & 7.54\%      \\
0.2       & 0.082          & 0.089          & 0.052          & 0.098          & 0.057          & 12.23\%      \\
0.3       & 0.121          & 0.096          & 0.056          & 0.101          & 0.059          & 15.65\%      \\
0.4       & 0.149          & 0.102          & 0.059          & 0.110          & 0.064          & 21.43\%      \\
0.5       & 0.165          & 0.124          & 0.072          & 0.127          & 0.074          & 23.21\%      \\
0.6       & \textbf{0.168} & \textbf{0.174} & \textbf{0.101} & \textbf{0.176} & \textbf{0.102} & 26.12\%      \\
0.7       & 0.162          & 0.152          & 0.089          & 0.155          & 0.090          & 32.76\%      \\
0.8       & 0.122         & 0.106          & 0.062          & 0.113          & 0.066          & 58.48\%      \\
0.9       & 0.080         & 0.043          & 0.025          & 0.052          & 0.030          & 79.76\%      \\
0.95      & 0.029         & 0.021          & 0.012          & 0.022          & 0.013          & 96.22\%      \\
0.999     & 0             & 0.005          & 0.003          & 0.007          & 0.004          & 98.30\%      \\ \hline
\end{tabular}
}
\caption{The detection and localization performance of the VAE anomaly detection method across different thresholds.}
\label{tab:vae_localization}
\end{table}

\paragraph{Comparison of the three methods.}
When comparing the results from the three methods (Table~\ref{tab:comparison_methods}, there are clear winners for the individual detection and localization tasks.
For detection, the classification method performs best with an F1-score of 0.791, followed by the object detector (0.720).
This is somewhat surprising as the object detector has been trained with most information (presence labels and location information).
For localization, the object detector seems to be the clear winner, which is expected giving the explicit location labels used for training.
However, during the qualitative analysis it was observed that the classification model almost exclusively has high activation around solar panels, despite being not so pixel-accurate as compare to the object detector.
Across all metrics, the anomaly detection performs worst and during qualitative analysis the anomaly detection seemed like a less focused version of the classification method, with abundant false positives in both detection and localization.

\begin{table}[!b]
\scalebox{0.75}{
\begin{tabular}{c|cc|cccc}
\hline
Method         & \multicolumn{2}{c|}{Detection}              & \multicolumn{4}{c}{Localization}       \\
               & Thrs.                      & F1-score        & Thrs.                     & DICE            & IoU            & No Overlap      \\ \hline
Object det.    & \multicolumn{1}{c|}{0.35}  & 0.720          & \multicolumn{1}{c|}{0.35} & \textbf{0.722}  & \textbf{0.611} & 1.12\%          \\
Classification & \multicolumn{1}{c|}{0.975} & \textbf{0.791} & \multicolumn{1}{c|}{0.7}  & 0.390           & 0.255          & \textbf{0.83\%} \\
Anomaly det.   & \multicolumn{1}{c|}{0.6}   & 0.168          & \multicolumn{1}{c|}{0.6}  & 0.174           & 0.101          & 26.12\%         \\ \hline
\end{tabular}
}
\caption{Scores for different methods by their best F1-score (detection) and their best DICE-score (localization).}
\label{tab:comparison_methods}
\end{table}



\subsection{Experiment 2: symmetry and asymmetry}

We measure the amount of symmetry on predictions between the different methods to investigate whether they succeed and fail on similar inputs.
On predictions that are asymmetrical (i.e. one method has a different answer than another), we perform a qualitative analysis to ascertain whether there are constant properties present in the input that lead to asymmetrical predictions among methods.
The results are shown in Table \ref{tab:symmetry}.

\begin{table}[!h]
\centering
\scalebox{0.75}{
\begin{tabular}{c|ccc}
\hline
\multicolumn{1}{c|}{$\downarrow$ Wrong \textbackslash  Right $\rightarrow$} & Object det. & Classification & Anomaly det. \\ \hline
Object detection             & 0                & 367            & 122               \\
Classification               & 217              & 0              & 91                \\
Anomaly detection            & 4,744            & 4,789          & 0                 \\ \hline
\#Right                      & 25,897           & 26,086         & 21,762            \\
\#Wrong                      & 746              & 557            & 4,881             \\ \hline
\end{tabular}}
\vspace{3mm}
\caption{Overview of quantitative symmetry between the methods.}
\vspace{3mm}
\label{tab:symmetry}
\end{table}

During qualitative analysis, the object detection and classification methods generally made similar mistakes.
These methods diverged on some points: we noticed that the object detection method was more prone to mistaking rectangular shapes on asphalt and concrete for solar panels and the classification method was more prone to mistaking rectangular shadows for solar panels.
As can also be seen in Table~\ref{tab:symmetry}, the anomaly detection method made many more mistakes compared to the object detection and classification methods.
Congruently, there are many properties on which the anomaly detection fails where the other methods succeed. 
One property that stood out was swimming pools: the anomaly detection method essentially always found swimming pools as anomalies.
As can be seen in Table \ref{tab:symmetry}, most of the mistakes of the anomaly detector are actually done right by the other two methods.
Therefore, the general performance might benefit from fused decisions from the object detection and classification methods.

\subsection{Experiment 3: decreasing amount of data}

\begin{figure}[!t]%
    \centering
    \includesvg[width=\linewidth]{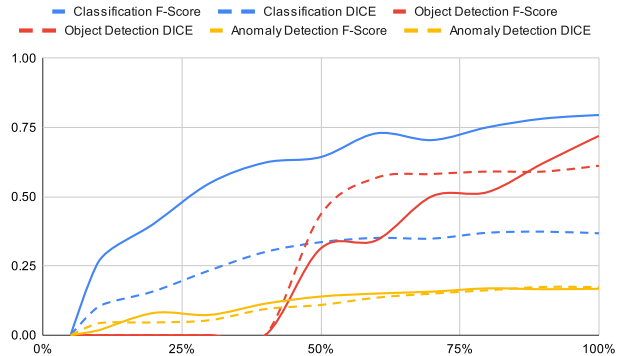}
    \caption{Detection and localization performance for all three methods when training with limited dataset size. Detection scores shown as solid lines and localization scores as dashed lines. Note that the classifier is the only method that performs reasonably well with limited data.}
    \label{fig:data_decrease_curve}
\end{figure}

Although the training set size is already limited, we evaluate the effect of limiting the training data size.
We decrease the size from 100\% in steps of 10\%, down to using only 1\% of the data.
The rest of the hyperparameters will remain the same, so by using 1\% of the data, while still keeping the same amount of training epochs, the model will observe fewer training samples.
As this has an effect on accuracy, we will investigate this as future work.
We use the optimal thresholds as determined in the previous experiments: object detection 0.35, classification (detection) 0.95, classification (localization) 0.7, anomaly detection (detection) 0.6, anomaly detection (localization) 0.6.
The results can be found in Figure~\ref{fig:data_decrease_curve} and Table~\ref{tab:data_decrease}.
Interesting is that across three separate runs, the object detection model never started giving correct output until using at least 40\% of the training set.
Especially when considering that at 40\% data usage the classification model already attained an F1-score of 0.643 and the anomaly detector model attained 0.140.
Furthermore, there is an increase in both detection and localization metrics when using more data across all models.
Interestingly, the localization performance seems to converge slightly earlier than the detection metrics. 
For all models it is the case that the more data we use, the better the performance becomes for both detection and localization.

\begin{table}[!t]
\centering
\scalebox{0.75}{
\begin{tabular}{c|cc|cc|cc}
\hline
           & \multicolumn{2}{c|}{Object Detection} & \multicolumn{2}{c|}{Classification}  & \multicolumn{2}{c}{Anomaly Detection} \\
Data Usage & F1-score & DICE  & F1-score & DICE  & F1-score & DICE  \\ \hline
1\%        & 0       & 0     & 0       & 0     & 0       & 0     \\
5\%        & 0       & 0     & 0.265   & 0.103 & 0.018   & 0.043 \\
10\%       & 0       & 0     & 0.403   & 0.159 & 0.081   & 0.046 \\
20\%       & 0       & 0     & 0.549   & 0.234 & 0.073   & 0.054 \\
30\%       & 0       & 0     & 0.624   & 0.301 & 0.114   & 0.094 \\
40\%       & 0.314   & 0.437 & 0.643   & 0.336 & 0.140   & 0.109 \\
50\%       & 0.343   & 0.569 & 0.729   & 0.351 & 0.150   & 0.136 \\
60\%       & 0.502   & 0.582 & 0.705   & 0.349 & 0.158   & 0.150 \\
70\%       & 0.517   & 0.591 & 0.751   & 0.370 & 0.170   & 0.162 \\
80\%       & 0.622   & 0.591 & 0.782   & 0.374 & 0.166   & 0.174 \\
90\%       & 0.720   & 0.612 & 0.795   & 0.368 & 0.168   & 0.173 \\
100\%      & 0.720   & 0.722 & 0.790   & 0.377 & 0.168   & 0.174 \\ \hline
\end{tabular}
}
\vspace{3mm}
\caption{The performance of different methods when using a subset of the training set.}
\vspace{3mm}
\label{tab:data_decrease}
\end{table}

\vspace{7mm}
\vspace{7mm}

\vspace{15mm}
\subsection{Training and evaluation time}
\label{sec:time}

We report the training and evaluation time for all models.
Training has been performed on a single GeForce RTX 3090 and testing on a single GeForce 1050 Ti.
Results are reported in Table~\ref{tab:time_measurements}.

\begin{table}[!t]
\centering
\scalebox{1.0}{
\begin{tabular}{l|r|r}
\hline
Model                          & Train time  & Test time  \\ \hline
Detection                      & 42m57s      & 14m26s \\
Classification + GradCAM       &  9m34s      &  4m38s \\
Classification + GradCAM++     &  9m34s      &  4m39s \\
Classification + HiResCAM      &  9m34s      & 12m33s \\
Classification + EigenCAM      &  9m34s      & 38m53s \\
Classification + EigenGradCAM  &  9m34s      & 43m16s \\
Classification + FullGrad      &  9m34s      & 53m21s \\
Anomaly detection              & 52m41s      & 17m35s\\ \hline
\end{tabular}
}
\caption{The training time per epoch and testing time for full test set evaluation (wall-clock time).}
\label{tab:time_measurements}
\end{table}

\vspace{5mm}
\section{Conclusions}
\label{sec:conclusions}

In this paper we have explored different levels of supervision and its effect on accuracy of presence detection and localization of objects in remote sensing imagery.
We have experimentally validated this for solar panel recognition in satellite images.
Three models have been evaluated, each with a different level of supervision.
Firstly, fully-supervised object detection is trained with location labels.
Secondly, image classification is trained with weakly supervised data by providing only image-level binary presence labels.
Lastly, anomaly detection is even more weakly supervised as it is trained using only images without objects (no labels).
Although the classification and anomaly detection don't perform explicit localization, we added a CAM method to explain classifications by generating heatmap activations in the input image.

The image classifier performs best in the binary presence detection task (F1-score 0.791), followed by the object detector (0.720).
The object detector generates the most accurate localization results, although qualitative evaluation shows that the CAM-based localizations are good.
The anomaly detector performed worst in both tasks and we expect that significantly more training data is required to obtain decent performance.
Evaluating the symmetry between the three models shows that accuracy can potentially be increased by fusing the model results.
The effect of the CAM method on the localization performance (for classifier and anomaly detector) is limited, and best results are obtained by GradCAM, GradCAM++ and HiResCAM, but the latter is not desired because of its high compute times.

Training the models with less data shows that the classifier is quite robust and can still obtain decent performance, while the object detector scores drops significantly if trained with less data.
The anomaly detection can also still work with less data, but the accuracy is much lower in general.

In summary, it is possible to find and localize solar panels on remotely sensed images using weakly supervised methods by using a classification model with a CAM-method as the localizer.
While the localization becomes slightly more imprecise, there are multiple advantages to using it over an object detection model.
The biggest advantage is requiring less labelling per image as the location of the solar panel does not need to be labelled.
Similarly, the classification method also requires less total labelled images in order before it starts being able to produce accurate recognition.
In addition, the classifier with CAM method if faster to train and evaluate, as compared to the object detector.

\vspace{5mm}
{\small
\bibliographystyle{ieeetran}
\bibliography{bibtex}
}

\end{document}